\documentclass[letterpaper, 10 pt, conference]{ieeeconf}
\IEEEoverridecommandlockouts
\usepackage[bookmarks=true,hidelinks]{hyperref}
\usepackage[utf8]{inputenc}
\usepackage{color}
\usepackage{float}
\usepackage{graphicx}
\graphicspath{ {Figures/} }

\usepackage{cite}
\usepackage[euler]{textgreek}
\usepackage[T1]{fontenc}
\usepackage{amsmath}
\usepackage{amssymb}
\usepackage{breqn}

\usepackage{comment}
\usepackage{bm}
\usepackage{siunitx}
\usepackage{xcolor}
\usepackage[ruled,vlined]{algorithm2e}

\SetCommentSty{mycommfont}
\DeclareSIUnit\pixel{pixel}
\usepackage{comment}
\usepackage{orcidlink} 

\begin{document}
%


\title{\LARGE \bf Fluoroscopy-Constrained Magnetic Robot Control via Zernike-Based Field Modeling and Nonlinear MPC %
\thanks{This work was supported by the National Science Foundation’s Foundational Research in Robotics CAREER
program under award number 2144348.}
\thanks{$^1$Laboratory for Computational Sensing and Robotics, Johns Hopkins University, Baltimore, MD 21218, USA. Email: {\tt\small xchen254@jhu.edu}.}%

\thanks{$^2$Department of Mechanical Engineering, University of Maryland, College Park, MD 20742, USA. Email: {\tt\small yancy@umd.edu}.}%
\thanks{$^3$Division of Magnetic Manipulation and Particle Research, Weinberg Medical Physics, MD 20852, USA. Email: {\tt\small lamar.mair@gmail.com}.}%
\thanks{\textsuperscript{\textasteriskcentered}Corresponding author}}%


\author{Xinhao Chen$^{1}$\textsuperscript{\textasteriskcentered}\orcidlink{0009-0007-7381-4127}, Hongkun Yao$^{1}$, Anuruddha Bhattacharjee$^{1}$\orcidlink{0000-0002-4053-4029}, Suraj Raval$^{2}$\orcidlink{0000-0003-2889-6841},\\ Lamar O. Mair$^{3}$\orcidlink{0000-0001-9459-3932}, Yancy Diaz-Mercado$^{2}$\orcidlink{0000-0003-0288-0112}, Axel Krieger$^{1}$\orcidlink{0000-0001-8169-075X}}

%


\maketitle
\begin{abstract}
Magnetic actuation enables surgical robots to navigate complex anatomical pathways while reducing tissue trauma and improving surgical precision. However, clinical deployment is limited by the challenges of controlling such systems under fluoroscopic imaging, which provides low frame rate and noisy pose feedback. This paper presents a control framework that remains accurate and stable under such conditions by combining a nonlinear model predictive control (NMPC) framework that directly outputs coil currents, an analytically differentiable magnetic field model based on Zernike polynomials, and a Kalman filter to estimate the robot state. Experimental validation is conducted with two magnetic robots in a 3D-printed fluid workspace and a spine phantom replicating drug delivery in the epidural space. Results show the proposed control method remains highly accurate when feedback is downsampled to 3 Hz with added Gaussian noise ($\sigma$ = 2 mm), mimicking clinical fluoroscopy. In the spine phantom experiments, the proposed method successfully executed a drug delivery trajectory with a root mean square (RMS) position error of 1.18 mm while maintaining safe clearance from critical anatomical boundaries.

\end{abstract}


\begin{figure*}[!t]
  \centering
  \includegraphics[width=\textwidth]{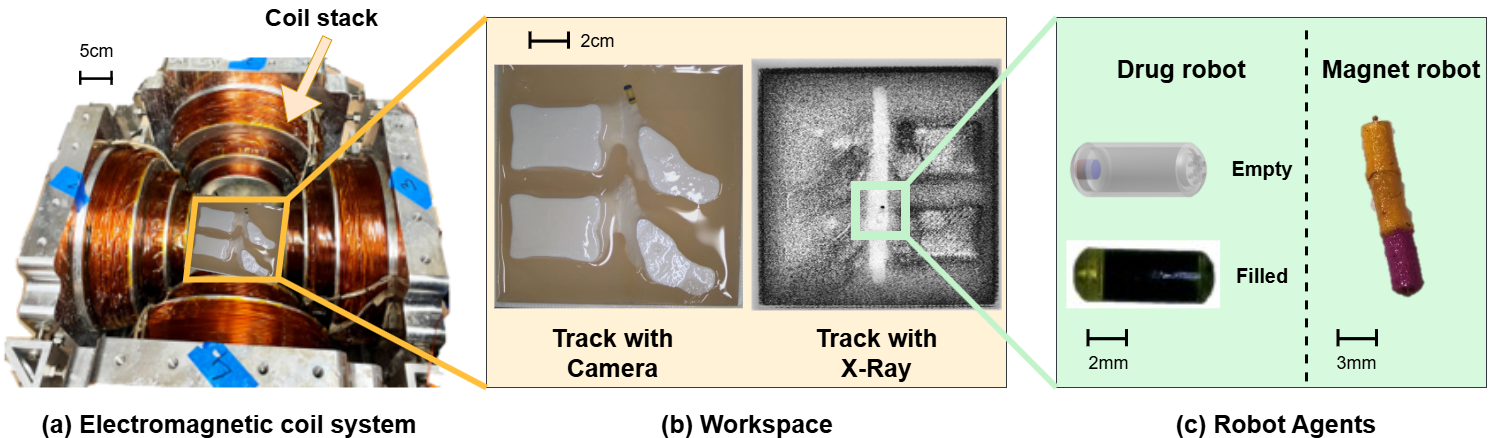}
  \vspace{-1em}
  \caption{(a) Electromagnetic coil system used in this study. The experimental workspace (yellow square) is centered between the coil stacks. (b) Spine phantom workspace imaged with an optical camera (left) and fluoroscopy X-rays (right). The robot agent is visible in both. (c) Robotic agents evaluated, including a drug-delivery capsule that can be filled with liquid medication.}
  \label{fig:system}
  \vspace{-1em}
\end{figure*}

\section{Introduction}
Magnetic actuation represents a transformative technology for minimally invasive surgical procedures~\cite{sliker2015magnetically}. Untethered magnetic systems allow navigating superficial and deep anatomical regions while minimizing tissue trauma by eliminating physical linkages to the end effector~\cite{sitti2015biomedical}. Emerging magnetic navigation approaches, ranging from externally steered microcatheters to microrobotic swarms, enable precise navigation and localized release under controlled magnetic fields, offering a promising route for focal therapy in the spinal canal~\cite{torlakcik2021magnetically} or other anatomical locations. 


Effective control is essential to map the motion of magnetically actuated robots to the input currents of electromagnetic coils. In recent years, learning-based control methods have been explored for this purpose~\cite{erin2022enhanced,cai2022deep,mao2025deep}. These approaches do not require explicit dynamics modeling of the robots and their interactions with the environment~\cite{li2024automated}. Examples include deep reinforcement learning for microrobots in fluidic tracks~\cite{behrens2022smart}, model-free reinforcement learning for a disk-shaped microrobot~\cite{salehi2024intelligent}, and data-driven optimal integrated controllers to reduce tracking error~\cite{wang2024data}. However, these methods often rely on large training datasets, operate at relatively low control update rates, and offer limited interpretability~\cite{li2024automated}.

In contrast, model-based methods typically follow a two-layer control architecture: a high-level controller computes the desired forces/torques or magnetic field/gradient at the robot's location, while a current allocator converts these quantities into coil currents using pre-calibrated unit-current field maps that vary with position in the workspace~\cite{abbott2020magnetic}. Systems such as OctoMag established this architecture~\cite{kummer2010octomag}, and subsequent work has largely refined the associated models and assumptions. 

At the controller layer, studies have used proportional–integral–derivative (PID) control~\cite{wang2025semi,tang2022vision}, introduced methods to mitigate actuation singularities~\cite{chen2024mitigating,raval2024singularity} or improve robustness~\cite{tang2022vision}, and applied sliding-mode control to handle disturbances~\cite{boroujeni2023five}. 
Few works have explored optimization-based control methods for improved performance under disturbances and uncertainties. Xu et al. deployed a linear quadratic regulator (LQR) controller to enable precise path-following with submillimeter accuracy~\cite{xu2021discrete}, while model predictive control (MPC) approaches incorporate look-ahead and constraint handling~\cite{abbott2020magnetic,he20253d,yang2021autonomous}.


To determine the coil currents, models map coil currents to the magnetic field and its spatial derivatives throughout the workspace. Modeling strategies exhibit diversity in their approaches and trade-offs. Simple analytical models, including point-dipole models and Biot-Savart law, offer high computational efficiency but limited accuracy~\cite{caciagli2018exact}. Complex analytical models, such as multipole or spherical-harmonic representations, require an initial calibration effort but subsequently provide improved accuracy~\cite{charreyron2021modeling}. Numerical models achieve the highest fidelity~\cite{bastos2003electromagnetic} but typically rely on interpolating unit-current lookup tables at runtime, which present challenges for gradient-based optimization~\cite{erin2022enhanced}. 
To address the challenge, Zernike polynomials can be used to fit the magnetic field data, providing an analytical formulation that maintains high accuracy while enabling efficient gradient computation~\cite{raval2021magnetic}. This study employs Zernike polynomials to approximate the high-fidelity numerical model.

Numerous surgical procedures implement fluoroscopic X-ray imaging to generate real-time visualization of internal anatomical structures, typically for verifying the placement of surgical tools~\cite{yamane2018effectiveness} as well as tracking magnetically controlled robotic agents. However, fluoroscopic tracking presents inherent limitations, including noisy feedback~\cite{uneri2017intraoperative} and relatively low frame rates (typically 2–15 frames per second~\cite{dias2016effects}) during the procedure~\cite{yamane2018effectiveness}. These constraints adversely affect the performance of existing control methods, which generally assume high-frequency, accurate (low noise) pose updates. Developing a new control method is therefore essential for achieving high-precision manipulation under such imaging conditions.

To render control methods robust to low frame rate fluoroscopic imaging, this paper introduces a framework specifically designed to operate under low frame rate and noisy tracking feedback. The primary contributions of this paper are (i) development of a Zernike polynomial-based model that provides a smooth, analytically differentiable representation of the magnetic field and its gradient across the workspace; (ii) a novel control framework based on nonlinear model predictive control (NMPC) and Kalman filter that reaches high control accuracy despite low frame rate, noisy tracking feedback, and (iii) experimental validation through phantom studies demonstrating the potential for targeted drug delivery within the spinal anatomy.

\section{Methods} \label{sec:Methods}

\subsection{Electromagnetic System and Magnetic Agents}
 
The electromagnetic actuation system employed in this study is illustrated in Fig.~\ref{fig:system}(a). This system, identical in configuration to that described in~\cite{erin2024strong}, comprises four coil stacks positioned on the lateral faces. Each stack consists of three independently controlled coils of varying dimensions. For this study, only the small and large coils were utilized, with each coil capable of operating at a maximum current of 30 A. The workspace, defined as a cubic volume with 100 mm edge length, is enclosed by the coil stacks.

Two robotic agents are employed in different experimental groups: a magnet robot and a drug delivery robot, as illustrated in Fig.~\ref{fig:system}(c). 
The magnet robot, operating in an unobstructed 10~cm $\times$ 10~cm workspace, features a cylindrical geometry with dimensions of 16 mm in length and 2.8 mm in diameter. It incorporates multiple NdFeB internal magnets, yielding a total magnetic moment of $1.35 \times 10^{-3}$ A$\cdot$m$^2$ oriented along its longitudinal axis. The main body of the robot was 3D printed using ABS (acrylonitrile butadiene styrene) material and subsequently painted pink and yellow to facilitate optical tracking. 
The drug delivery robot is specifically designed for intravascular or intraspinal applications. This cylindrical device measures 7.4 mm in length and 2.8 mm in diameter, incorporating a hollow chamber with a capacity of approximately 21.6 mm$^3$ for liquid drug storage. One end of the drug delivery robot has seven small holes that enable both drug loading and passive diffusion. This design ensures minimal drug release during transit, with the majority of the drug solution released to the target location through sustained positioning beside the target. The non-perforated end contains a neodymium grade N50 disk magnet (1 mm diameter, 1 mm length), generating a magnetic moment of approximately $8.9 \times 10^{-4}$ A$\cdot$m$^2$. The polymeric components were made using biocompatible resin on a microArch\textsuperscript{\textregistered} S140 3D printer.

\begin{figure}
\centering
\includegraphics[width=0.98\linewidth]{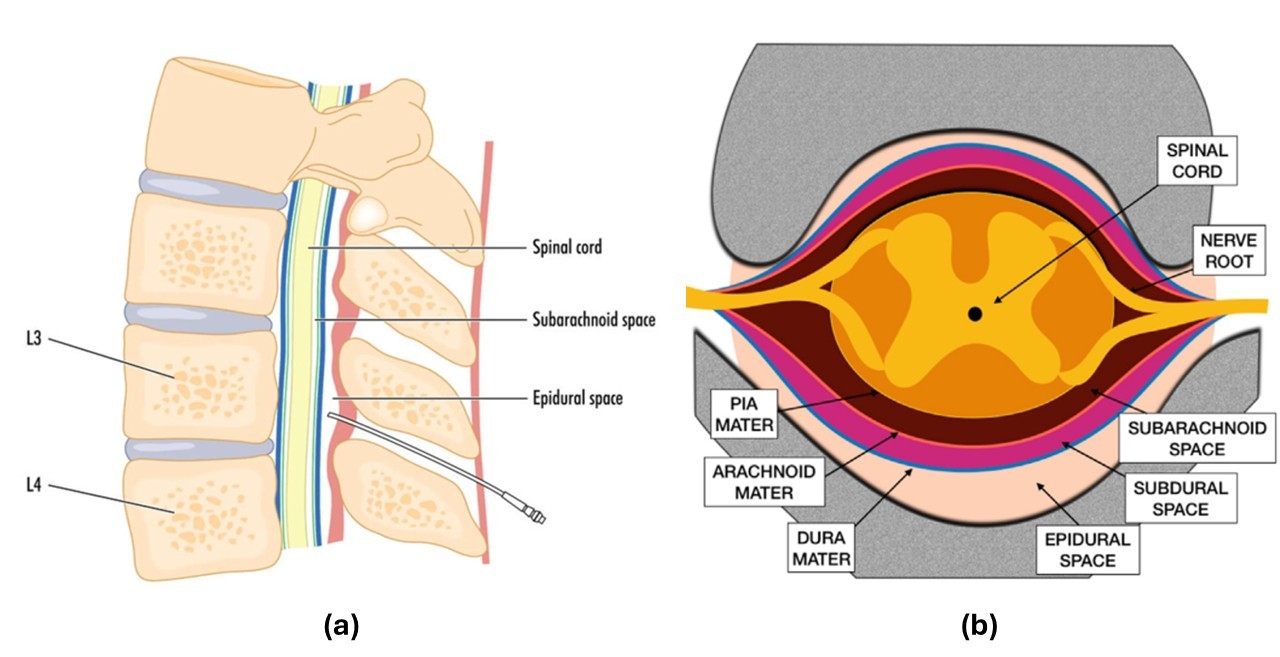}
\vspace{-1em}
\caption{(a) The spine structure in sagittal (longitudinal) plane.
(b) The spine structure in transverse (horizontal) plane.~\cite{nagel2018spinal}\label{fig:spine}}
\vspace{-1em}
\end{figure}

\begin{figure}
\centering
\includegraphics[width=0.98\linewidth]{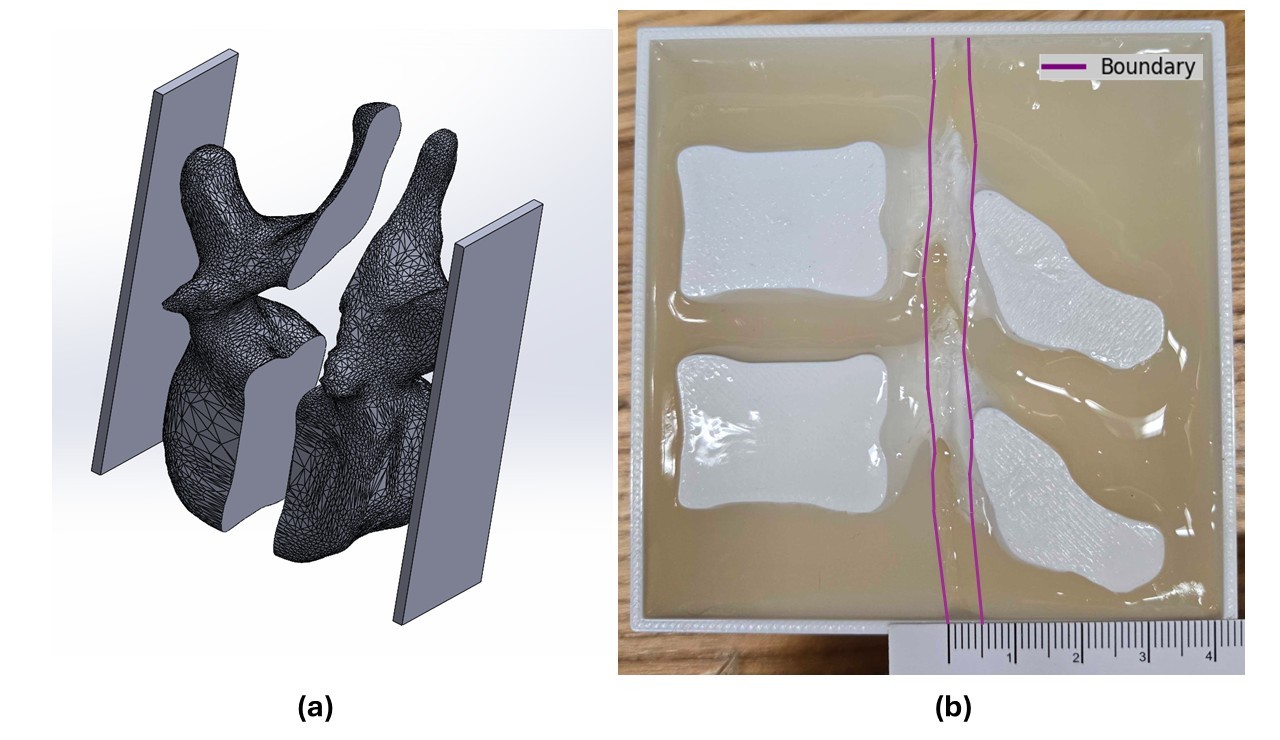}
\vspace{-1em}
\caption{(a) CAD model of a human vertebral segment sectioned in the sagittal plane. (b) Spine phantom fabricated from ABS, agarose gel, and a glycerin–water solution. The epidural space is indicated by purple lines.\label{fig:med_workspace}}
\vspace{-1em}
\end{figure}

Two experimental workspaces were developed for this study: a 3D-printed workspace and a spine phantom workspace. The 3D-printed workspace is an open container filled with a 50\% glycerin-water solution. This concentration was selected to match the viscosity of human blood~\cite{ekbal2013monitoring}. The workspace was printed using ABS material on a Stratasys F370 printer.
The spine phantom workspace was constructed within a 3D-printed enclosure to replicate the conditions for intraspinal drug delivery. As illustrated in Fig.~\ref{fig:spine}, the epidural space represents one anatomical region where surgical instruments and robotic agents can safely operate. This space exhibits an anteroposterior thickness of 4 to 7 mm in the sagittal plane~\cite{gala2016imaging} and presents a characteristic U-shaped configuration in the transverse plane. To construct the phantom, as shown in Fig.~\ref{fig:med_workspace}(a), human spine models of the L3/L4 vertebrae were sectioned along the sagittal plane and printed with supporting structures. Two adjacent sections were positioned within the enclosure, and the remaining volume was filled with 1.2\% agarose gel solution. Following gel solidification, a 5 mm wide channel was carved to simulate the epidural space, indicated by purple lines in Fig.~\ref{fig:med_workspace}(b). The channel was filled with a 10\% glycerin-water solution, selected to match the viscosity characteristics of cerebrospinal fluid (CSF)~\cite{bloomfield1998effects}.

\subsection{Zernike-Based Magnetic Field \& Gradient Modeling}

Although a well-established finite element analysis (FEA) model with high accuracy exists for this system~\cite{chen2024mitigating}, it relies on interpolating lookup tables during operation. Thus, an alternative model offering comparable accuracy while providing analytic expressions and derivatives is sought.

Zernike polynomials constitute a complete orthogonal basis on the unit disk and were extensively employed in optical testing and wavefront analysis owing to their compact representation and physical interpretability~\cite{WyantZernike, raval2021magnetic}. Zernike polynomials make an ideal basis to represent magnetic fields due to their polar-coordinates symmetry. In polar coordinates $(\rho,\varphi)$, with radial distance $0 \leq \rho \leq 1$ and azimuthal angle $0 \leq \varphi < 2\pi$, the polynomials are conventionally separated into two families distinguished by their angular symmetry.
The even Zernike polynomials are

\begin{equation}
Z_n^{m}(\rho,\varphi) = R_n^m(\rho)\cos(m\varphi),
\end{equation}
while the odd Zernike polynomials are
\begin{equation}
Z_n^{-m}(\rho,\varphi) = R_n^m(\rho)\sin(m\varphi).
\end{equation}

\noindent where $n$ denotes the radial degree and $m$ the azimuthal frequency, and they are non-negative integers with ${n \geq m \geq 0}$.
The radial functions $R_n^m(\rho)$ are defined by
\begin{multline}
R_n^m(\rho) = \\
\sum_{k=0}^{(n-m)/2}
  \frac{(-1)^k (n-k)!}{k!\,\left(\tfrac{n+m}{2}-k\right)!\,\left(\tfrac{n-m}{2}-k\right)!}
  \rho^{\,n-2k}.
\end{multline}

Since $Z_n^m(x,y)$ are polynomial functions of $(x,y)$, 
their derivatives $\partial Z_n^m / \partial x$ and $\partial Z_n^m / \partial y$ 
can be computed analytically. This property enables direct calculation of 
magnetic field gradients $\nabla \mathbf{B}$, which are essential for estimating 
forces and torques on the robot without relying on numerical differencing, which suffers under noisy measurements.

While Zernike functions are typically expressed in polar form, we adopt their Cartesian polynomial expressions 
as tabulated in \cite{WyantZernike} for practical implementation in our system.
Table~\ref{tab:zernike_cartesian} lists several low-order Zernike polynomials, 
which illustrate how the basis functions reduce to simple monomials in $x$ and $y$. 
\begin{table}[ht]
\centering
\vspace{-0.5em}
\caption{Zernike polynomials in Cartesian coordinates \cite{WyantZernike}.}
\label{tab:zernike_cartesian}
\begin{tabular}{c c c l}
\hline
\# & $n$ & $m$ & Polynomial \\
\hline
0 & 0 & 0 & $1$ \\
1 & 1 & 1 & $x$ \\
2 & 1 & 1 & $y$ \\
3 & 2 & 0 & $2(x^2+y^2)-1$ \\
4 & 2 & 2 & $x^2 - y^2$ \\
5 & 2 & 2 & $2xy$ \\
\hline
\end{tabular}
\vspace{-0.5em}
\end{table}

FEA-based modeling was employed for this system to enhance the 
estimation accuracy of near-source magnetic fields. Magnetic field data $\{B_x(x_i,y_i),\,B_y(x_i,y_i)\}$ were generated in COMSOL for three 
coil configurations (small, medium, and large)~\cite{chen2024mitigating}. A magnetic field was then fit with Zernike polynomials on scaled and shifted disks centered at each one of the coils. The coil centers relative to the workspace 
origin and the corresponding disk radii that can cover the whole workspace are summarized in 
Table~\ref{tab:coil_params}. To obtain a compact and differentiable 
representation, the distribution is here approximated by a truncated Zernike expansion. 
The in-plane components are expressed as
\begin{multline}
\label{eq:Zernike}
B_x(x,y) \approx \sum_{n,m} a_{n,m}\, Z_n^m(x,y),\\
B_y(x,y) \approx \sum_{n,m} b_{n,m}\, Z_n^m(x,y)
\end{multline}
with coefficients $\{a_{n,m}\}$ and $\{b_{n,m}\}$ identified by least-squares fitting.

\begin{table}[ht]
\centering
\vspace{-0.5em}
\caption{Coil centers and normalization radii
}
\label{tab:coil_params}
\begin{tabular}{lcc}
\hline
Configuration & Coil center (mm) & $\rho_{\max}$ (mm) \\
\hline
Small  & $(0,\,57.69)$  & $118.731$ \\
Medium & $(0,\,71.295)$ & $136.186$ \\
Large  & $(0,\,90.925)$ & $165.341$ \\
\hline
\end{tabular}
\vspace{-0.5em}
\end{table}



\begin{figure}
\centering
\includegraphics[width=0.98\linewidth]{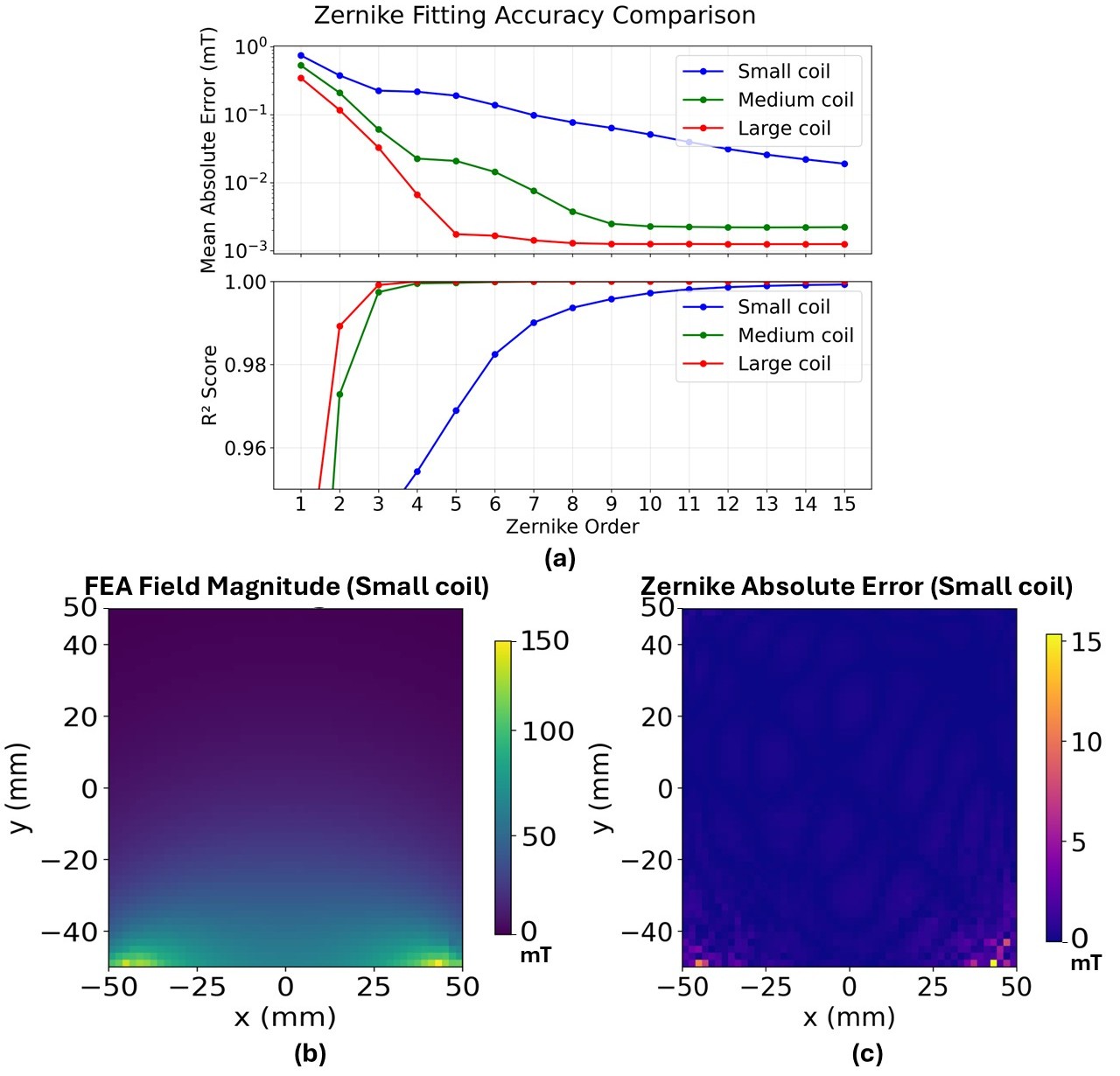}
\vspace{-1em}
\caption{ (a) Mean absolute error (MAE) and $R^2$ score of Zernike model with different polynomial order. (b) FEA-computed magnetic-field magnitude for a small coil at unit current (coil located at the bottom of the figure). (c) Absolute fitting-error distribution for the Zernike model under the same condition.
\label{fig:zernike}}
\vspace{-1em}
\end{figure}

The truncation order $N$ was selected by evaluating fitting accuracy using two metrics over the sampled domain: mean absolute error (MAE) and the
coefficient of determination ($R^2$), shown in Fig.~\ref{fig:zernike}(a). For the medium and large coils, order $N=3$ is selected as $R^2$
saturates by that point, while the small coil uses order $N=4$, reflecting its higher spatial-frequency content near the coil edges.

To illustrate reconstruction quality, Fig.~\ref{fig:zernike}(b) shows the FEA magnetic
field magnitude for a small coil at unit current, and Fig.~\ref{fig:zernike}(c) depicts the absolute
error of the corresponding Zernike model. 
The fitting errors are confined to below 4 mT across the majority of the workspace, increasing only in regions near coil boundaries characterized by steep field gradients. This level of accuracy satisfies the requirements for effective magnetic control. Furthermore, the functions were confirmed to be divergence-free, consistent with Gauss’s law for magnetism.


\subsection{NMPC and Kalman Filter Based Control}


Unlike conventional two-layer architectures that first compute desired forces/torques followed by input currents, the proposed controller employs a unified NMPC framework that directly maps robot and desired poses to coil currents. 

\begin{figure}
\centering
\includegraphics[width=0.98\linewidth]{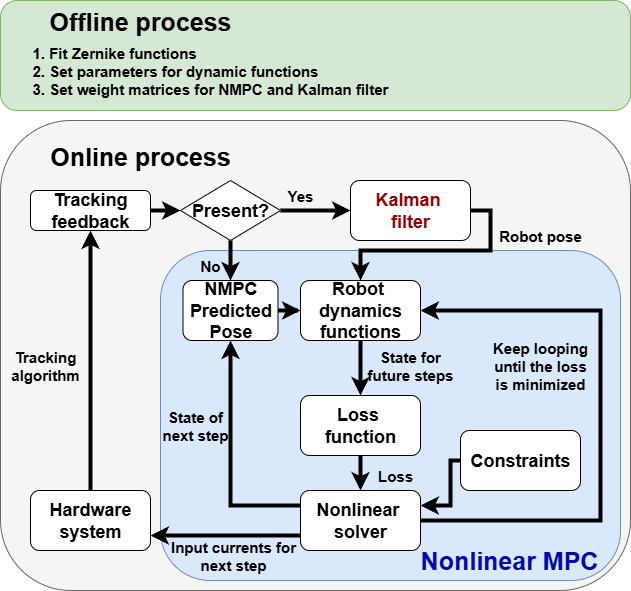}
\vspace{-1em}
\caption{ Flow chart of the proposed control method
\label{fig:MPC}}
\vspace{-1em}
\end{figure}

The controller architecture, depicted in Fig.~\ref{fig:MPC}, consists of offline and online components. Offline processes include Zernike polynomial fitting to the FEA model, dynamics parameter configuration, and setting the weight matrix for the NMPC and Kalman filter.

The online process begins by determining the robot pose input for the NMPC. When tracking feedback is available in the current control loop, the Kalman filter estimates the current robot pose by fusing the feedback with the NMPC-predicted pose from the previous control loop which serves as the state from dynamic model. When tracking feedback is unavailable due to its lower update rate relative to the control frequency, the NMPC-predicted pose is used directly as the current robot pose. The velocities are subsequently computed from the pose change over the time step.

The state of the robot inside NMPC is defined as
\begin{equation}
\label{eq:robot_state}
\mathbf{x} := [\,x,\ y,\ \theta,\ v_x,\ v_y,\ \omega\,]^\top
\end{equation}
where $(x,y)$ and $\theta$ are position and orientation and $(v_x,v_y,\omega)$ are their rates.

The basic equation of motion for a floating robot is given by
\begin{equation}
\label{eq:floating_robot}
\mathbf{M} \cdot \mathbf{\dot{v}}(t)+\mathbf{D}(\mathbf{v}(t))+\mathbf{C}(\mathbf{v}(t))=\mathbf{F}(t)
\end{equation}
where $\mathbf{M}$ represents the mass and inertia matrix, $\mathbf{D}(\mathbf{v}(t))$ denotes the hydrodynamic damping forces/torques, and $\mathbf{C}(\mathbf{v}(t))$ represents the Coriolis and centripetal forces.

In the context of this study, the Coriolis and centripetal forces are negligible. The hydrodynamic damping forces are simplified to the drag force and torque proportional to the robot's velocity. Consequently, Eq.~\eqref{eq:floating_robot} reduces to

\newcommand{\R}[1]{\mathbf{R}(#1)}
\begin{equation}
\label{eq:dynamics}
\begin{aligned}
&m\begin{bmatrix}\dot{v}_x\\ \dot{v}_y\end{bmatrix}
= \mathbf{F}_{\mathrm{mag}}
- \mathbf{D}_t(\theta)\begin{bmatrix}v_x\\ v_y\end{bmatrix}\\
&J\,\dot{\omega} = \tau_{\mathrm{mag}} - d_r\,\omega\\
&\mathbf{D}_t(\theta) = \R{\theta}\!\begin{bmatrix}d_{\parallel} & 0\\[2pt] 0 & d_{\perp}\end{bmatrix}\!\R{\theta}^\top
\end{aligned}
\end{equation}
where $m$ is the mass and $J$ is the rotational inertia. The parameters $d_{\parallel}$, $d_{\perp}$, and $d_r$ denote the drag coefficients. The rotation matrix $\mathbf{R}(\theta)$ transforms the drag force matrix from the body frame to the world frame. The magnetic force $\mathbf{F}_{\mathrm{mag}}$ and torque $\tau_{\mathrm{mag}}$ are computed using Eq.~\eqref{eq:magnetic_force}.

\begin{equation}
\label{eq:magnetic_force}
\mathbf{F}_{\mathrm{mag}}= (\boldsymbol{m}^T \cdot \nabla) \boldsymbol{B} =  \begin{bmatrix} \boldsymbol{m}^T  \frac{\partial \boldsymbol{B}}{\partial x} \\ \boldsymbol{m}^T  \frac{\partial \boldsymbol{B}}{\partial y}\end{bmatrix}, \quad
\tau_{\mathrm{mag}}= \boldsymbol{m} \times \boldsymbol{B}.
\end{equation}
where $\boldsymbol{m}$ is the robot's magnetic moment and $\boldsymbol{B}$ is the magnetic field vector. Both $\boldsymbol{B}$ and its spatial gradient are computed from the robot position and coil currents $\boldsymbol{I}$ based on the Zernike model. 

Eq.~\eqref{eq:Zernike}, \eqref{eq:dynamics}, and \eqref{eq:magnetic_force} establish the relationship between the coil current vector $\boldsymbol{I}$ and the robot state $\mathbf{x}$, providing the dynamics function required for NMPC implementation.

The NMPC formulation requires a cost function to minimize. For a prediction horizon $N$ with $k = 0, \ldots, N-1$, the robot pose errors are defined as
\begin{equation}
\begin{aligned}
e_{\mathbf{p},k}&=
\begin{bmatrix}
x_k-x_{desired}\\[2pt]
y_k-y_{desired}
\end{bmatrix},\\
e_{\theta,k} &= \bigl( \,(\theta_k - \theta_{desired} + \pi) \bmod 2\pi \,\bigr) - \pi
\end{aligned}
\end{equation}
with $\Delta\mathbf{I}_k=\mathbf{I}_k-\mathbf{I}_{k-1}$. The $e_{\theta,k}$ is calculating the difference between $\theta_k$ and $\theta_{desired}$ without the influence from $2\pi$ periodicity. The stage and terminal costs are
\begin{equation}
\begin{aligned}
\ell_k &= e_{\mathbf{p},k}^\top \mathbf{Q}_p\, e_{\mathbf{p},k}
+ q_\theta\, e_{\theta,k}^2
+ \mathbf{I}_k^\top \mathbf{R}\,\mathbf{I}_k
+ \Delta\mathbf{I}_k^\top \mathbf{S}\,\Delta\mathbf{I}_k,\\
\ell_N &= e_{\mathbf{p},N}^\top \mathbf{P}\, e_{\mathbf{p},N}
\;+\; q_{\theta,N}\, e_{\theta,N}^2
\end{aligned}
\end{equation}
The total loss minimized by NMPC is
\begin{equation}
Loss = \ell_N+\sum_{k=0}^{N-1} \ell_k 
\end{equation}
with positive semi-definite weights $\mathbf{Q}_p$, $q_\theta$, $\mathbf{R}$, $\mathbf{S}$, and $\mathbf{P}$. The loss function enforces pose tracking while regularizing current magnitude and current changing rate to ensure smooth, safe actuation. Through extensive tuning, it was observed that the control framework has the best performance when the norm of $\mathbf{R}$ is approximately two times that of $\mathbf{S}$, and the norm of $\mathbf{Q}_p$ is ten times that of $\mathbf{R}$.

At each control loop, the nonlinear optimal control problem is solved over a prediction horizon $N = 20$ with a time step of $\Delta t = 40$ ms,
using CasADi-based automatic differentiation and IPOPT as the nonlinear program solver. The decision variables are the coil current sequence and the predicted state trajectory. The constraints added to the solver include current limits, current changing rate limits, and workspace limits. Upon convergence, the first element of the optimal current sequence is applied to the hardware system, while the corresponding pose prediction is stored for robot pose estimation in the subsequent control loop.
 
The single NMPC architecture offers two key advantages over ``two-layer'' control approaches. First, it incorporates hardware constraints, whereas high-level controllers in two-layer methods may command forces or torques that exceed the system's physical capabilities. Second, by integrating current computation within the closed-loop optimization, the NMPC can compensate for magnetic field modeling errors, a capability absent in two-layer methods where current allocation occurs outside the feedback loop.

\subsection{Comparative Control Methods}

Six control methods are evaluated: the proposed framework and five baselines for comparison. The NMPC parameters across all baselines are identical to those of the proposed framework.

\textbf{Baseline 1:} Implements the identical NMPC framework as the proposed method but omits the Kalman filter, relying directly on camera-based pose measurements when available and switching to NMPC predictions during measurement gaps.

\textbf{Baseline 2:} Two-layer control architecture employing a well-tuned PID controller for force and torque computation from robot pose, with current calculation based on the FEA look-up tables.

\textbf{Baseline 3:} Two-layer control architecture employing a well-tuned linear MPC controller for force and torque computation from robot pose, with current calculation based on the FEA look-up tables. Employ MPC predictions when tracking feedback is not available.

\textbf{Baseline 4:} Two-layer control architecture identical to Baseline 3, except currents are calculated using the Zernike model.

\textbf{Baseline 5:} Modified version of the proposed control framework. Instead of the Zernike model, it incorporates FEA lookup tables, together with their precomputed Jacobian matrices providing gradient information, within the NMPC framework.

\subsection{Tracking Algorithm and Mimic Fluroscopy Feedback}

The workspace can be visualized from the top view using cameras or various medical imaging modalities, including C-arm X-ray fluoroscopy. Robotic agents can subsequently be localized through image-based tracking algorithms. Due to the electromagnetic actuation system's weight constraints, fluoroscopic tracking was not feasible for this study. Therefore, camera-based tracking was employed to simulate fluoroscopic feedback characteristics.

In this study, a FLIR BFS-U3-13Y3 camera mounted above the workspace provides visual feedback, with pose estimation performed using a modified UNet architecture adapted from prior work~\cite{pryor2021localization}. The training dataset comprises 572 × 572 RGB images with annotations. These annotations were generated by first segmenting the raw images using Segment Anything Model 2 (SAM2)~\cite{ravi2024sam}, then by manual identification of three fiducial points on each robot mask.

The camera-based tracking system demonstrates negligible error, significantly outperforming typical fluoroscopic tracking accuracy. Current fluoroscopic tracking methods for pedicle screws in spine surgery exhibit errors from 1.1 mm to 2.7 mm, varying with the specific technique and clinical scenario~\cite{uneri2017intraoperative}. Given that the robotic agents are smaller than pedicle screws, fluoroscopic tracking errors may be higher for them.

Furthermore, intraoperative fluoroscopy for spinal interventions typically operates at frame rates around 7.5 frames per second (fps)~\cite{yamane2018effectiveness}. During critical phases such as pedicle screw placement, systems are configured to 15 fps to maximize temporal resolution~\cite{bullert2025markerless}, while less complex procedures may employ reduced frame rates of 3 Hz to minimize radiation exposure~\cite{sabat2018radiation}. To maintain radiation exposure within acceptable limits, cumulative fluoroscopy time is not recommended to exceed 60 minutes~\cite{stecker2009guidelines}. While most advanced intraoperative fluoroscopy may operate at higher rates exceeding 15 fps, maintaining robot control at a lower frequency extends the allowable procedure duration and reduces total radiation dose.

The robotic agents' visibility under X-ray imaging was confirmed, as illustrated in Fig.~\ref{fig:system}(b). To simulate the characteristics of fluoroscopic tracking, the camera-based pose feedback was artificially degraded by reducing the frame rate and introducing Gaussian noise before transmission to the controller. Meanwhile, the raw camera-based pose feedback with high frame rate was used as the robot's actual pose to calculate the control errors towards the desired trajectory.

\section{Experiments and Results}\label{Sec:Results}

Five experiments were conducted to progressively validate the proposed control method's capability for fluoroscopy-guided drug delivery in spine: (1) validation of NMPC pose-prediction accuracy; (2-4) closed-loop control of the magnet robot in an unobstructed workspace under three experimental conditions; and (5) drug delivery demonstration in a spine phantom. All reported errors represent mean values from three independent trials per experimental condition.



\subsection{NMPC Pose Prediction Accuracy Test}

\begin{figure}
\centering
\includegraphics[width=0.98\linewidth]{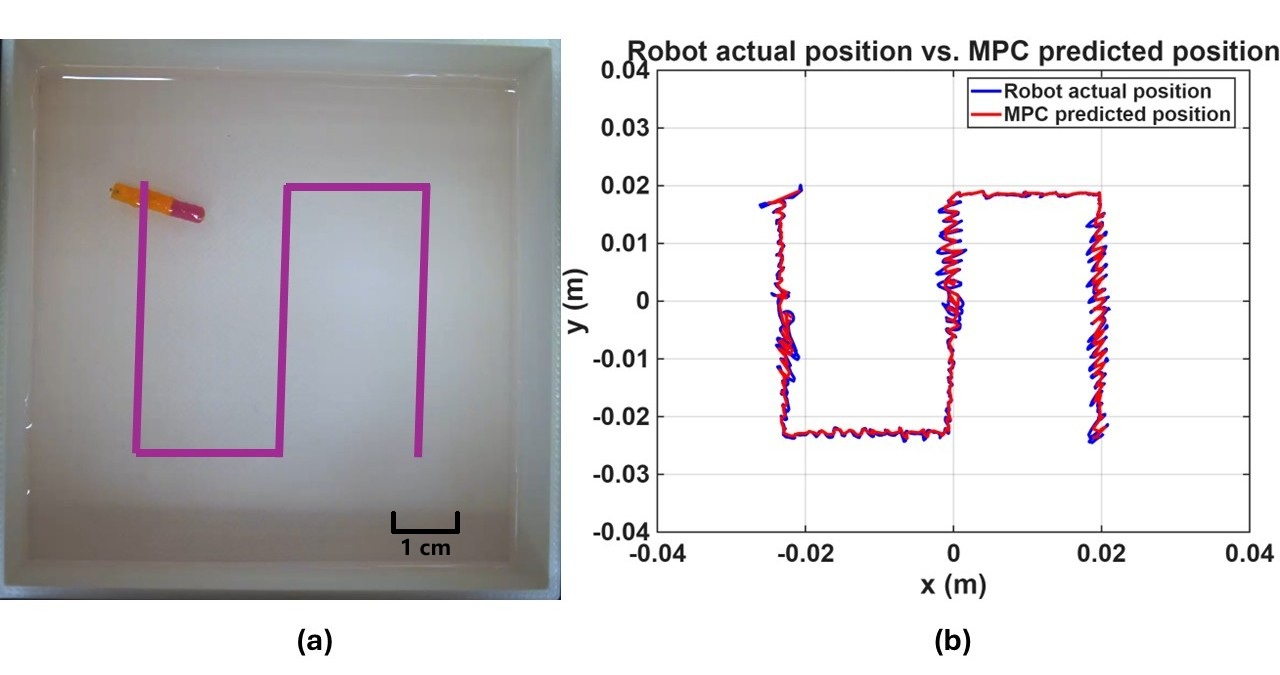}
\vspace{-1em}
\caption{(a) The workspace for the magnet robot. The purple line shows the desired trajectory.
(b) Robot actual and NMPC-predicted trajectories. \label{fig:MPC_prediction}}
\vspace{-1em}
\end{figure}

The magnet robot was controlled using high frame rate, accurate tracking feedback in the unobstructed workspace. An S-shaped trajectory, defined within a $43 \times 43$ mm square, was implemented for the magnet robot to ensure comprehensive workspace utilization while maintaining safe clearance from boundaries, as shown in Fig.~\ref{fig:MPC_prediction}(a). 

During each control loop, both robots' actual pose and NMPC-predicted pose from the previous loop were recorded. The root mean square (RMS) position error between actual and predicted positions was 0.69 mm, with an RMS orientation error of \SI{6.0}{\degree}. As illustrated in Fig.~\ref{fig:MPC_prediction}(b), the robot's actual trajectory (blue line) closely aligns with the NMPC-predicted trajectory (red line), demonstrating the reliability of the NMPC-predicted pose.

\subsection{Errors vs Tracking Rate}

The magnet robot was subsequently controlled using six different control methods under varying tracking feedback rates, ranging from 3 Hz (the minimum intraoperative fluoroscopy frame rate) to 25.5 Hz (the maximum achievable control rate across all methods).

Preliminary tests demonstrated that Baseline 5's computational demands made it impractical, as the NMPC optimization required excessive lookup table interpolations per control iteration. With control rates degrading below 0.5 Hz, this method was incapable to complete task and was therefore excluded from subsequent experiment. This result validates the necessity of the Zernike model.

\begin{figure}
\centering
\includegraphics[width=0.98\linewidth]{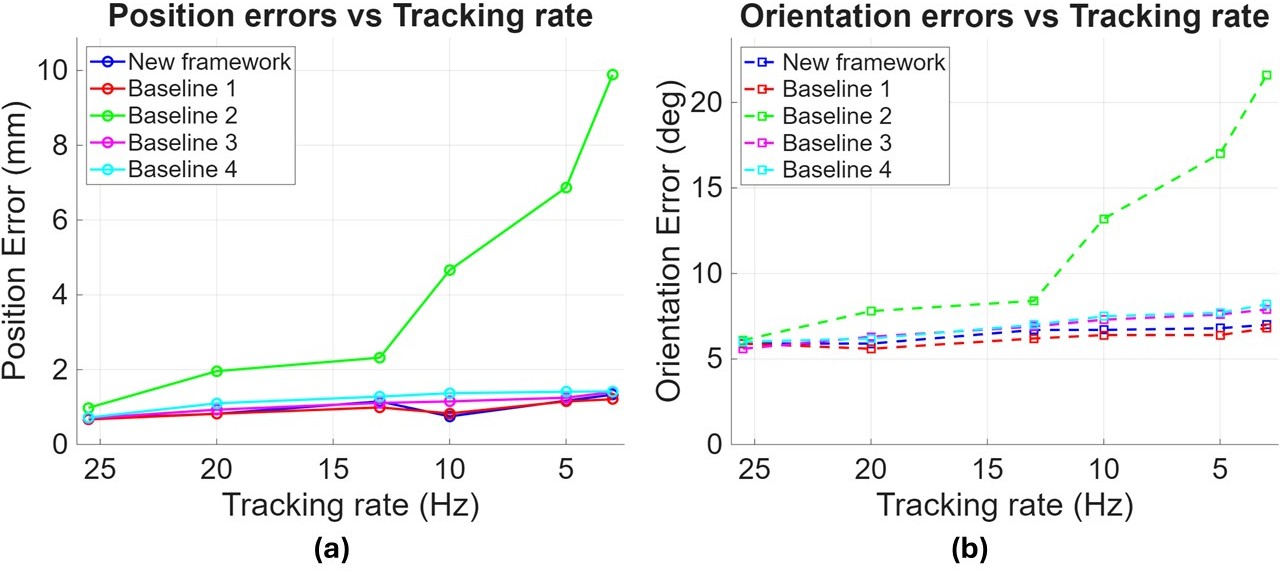}
\vspace{-1em}
\caption{(a) RMS position errors under different tracking rates.
(b) RMS orientation errors under different tracking rates. \label{fig:results_rate}}
\vspace{-1em}
\end{figure}

Fig.~\ref{fig:results_rate} presents the results for other methods. Notably, all of them achieved submillimeter accuracy with 25.5 Hz pose feedback, confirming that controller parameters are well-tuned. Except for Baseline 2 (PID controller), all methods maintain relatively stable performance when tracking rates decrease. This stability can be attributed to their reliance on NMPC/MPC-predicted poses, which prove reliable when pose feedback is accurate. The proposed controller exhibits marginally higher error than the best-performing method (by 0.05 mm), potentially due to the Kalman filter preventing full utilization of the accurate camera-based pose feedback.

\subsection{Errors vs Tracking Noise}

Gaussian noise is artificially introduced to the camera-based feedback, with noise level defined by the standard deviation $\sigma$ of the Gaussian distribution. The tracking feedback was provided at 25.5 Hz.

\begin{figure}[b]
\centering
\includegraphics[width=0.98\linewidth]{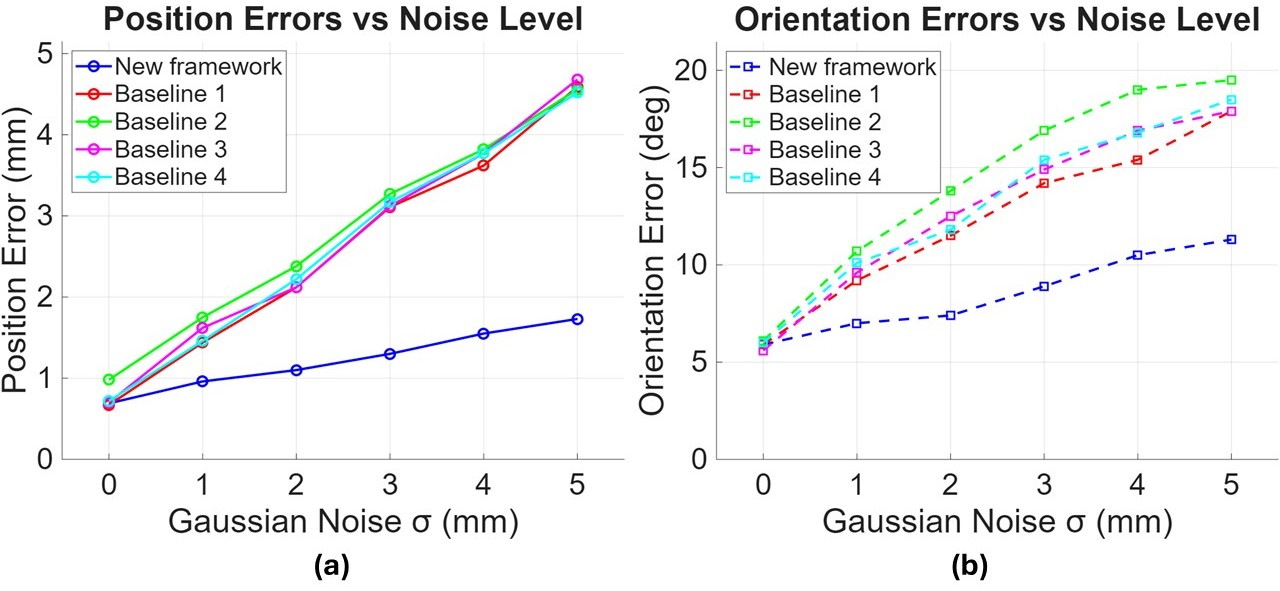}
\vspace{-0.5em}
\caption{(a) RMS Position errors under different tracking noise levels.
(b) RMS Orientation errors under different tracking noise levels. \label{fig:results_noise}}
\vspace{-0.5em}
\end{figure}

As shown in Fig.~\ref{fig:results_noise}, all baseline methods exhibit substantial degradation with rising noise levels, whereas the proposed control method demonstrates reduced degradation. This disparity indicates that NMPC/MPC-predicted poses become unreliable under noisy feedback conditions, as the NMPC/MPC are fed incorrect poses. In contrast, the Kalman filter effectively mitigates these noise-induced errors.

\subsection{Errors vs Tracking Rate Under Fixed Noise Level}

Gaussian noise with $\sigma = 2$ mm is added to the tracking feedback, chosen to match the noise characteristics of intraoperative fluoroscopy. Robot performance was evaluated under varying tracking rates to simulate fluoroscopy operating at different frame rates during surgical procedures.

\begin{figure}[b]
\centering
\includegraphics[width=0.98\linewidth]{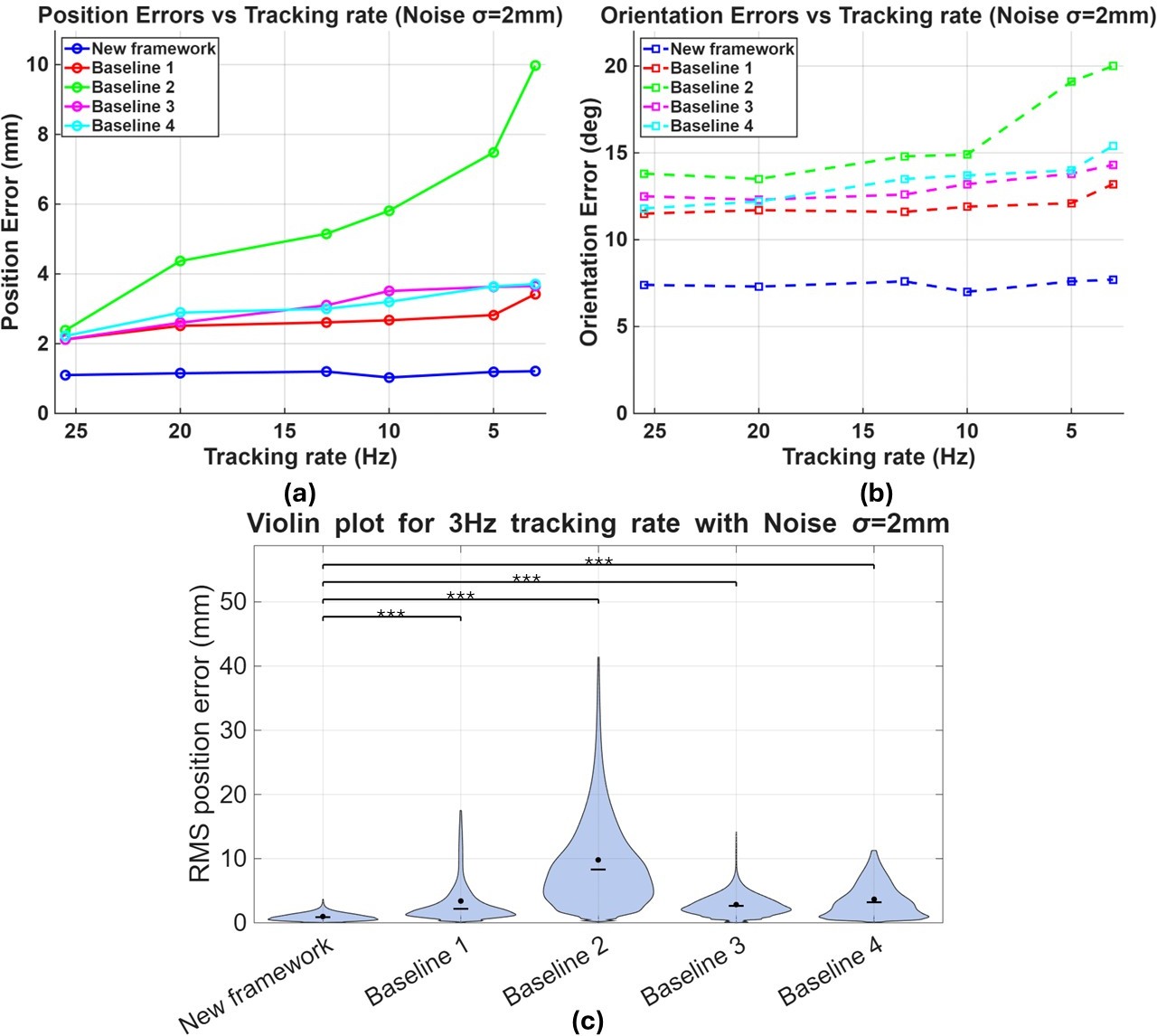}
\vspace{-0.5em}
\caption{(a) RMS position errors under different tracking rates with fixed noise level.
(b) RMS orientation errors under different tracking rates with fixed noise level. 
(c) Violin plot at 3Hz with fixed noise level. All tests has $p < 10^{-3}$, denoted by ***.\label{fig:results_mimic}}
\vspace{-0.5em}
\end{figure}

As shown in Fig.~\ref{fig:results_mimic}, Baselines 1, 3, and 4 maintain relatively stable performance as tracking rates decrease, while Baseline 2 exhibits dramatic error increases. All P values in the violin plot are smaller than $10^{-3}$, which shows the evidence that the proposed method demonstrates superior performance. These results effectively combined the trends observed in the previous two experimental groups.

\subsection{Spine Phantom Experiment}

\begin{figure}
\centering
\includegraphics[width=0.98\linewidth]{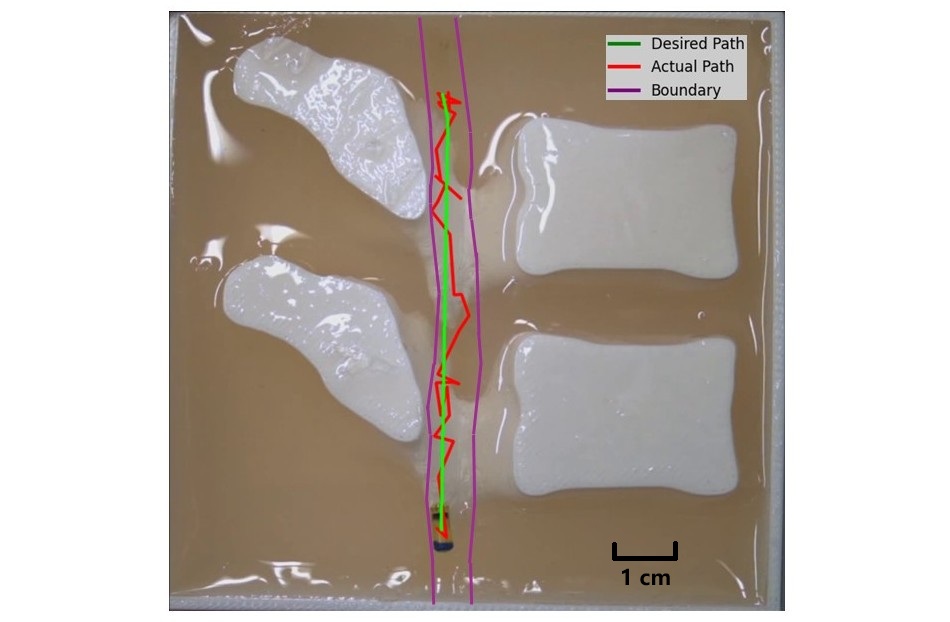}
\vspace{-0.5em}
\caption{Results of spine-phantom experiment. The green curve denotes the desired trajectory; the red curve shows the actual robot trajectory; purple lines delineate the epidural space's boundaries. \label{fig:results_med}}
\vspace{-0.5em}
\end{figure}

The final experiment employed the spine phantom under challenging conditions: 3 Hz tracking rate (minimum fluoroscopy frame rate) with added Gaussian noise ($\sigma = 2$ mm). Given the degraded performance of baseline methods under this condition, only the proposed controller was tested.

The trajectory was designed to navigate within the epidural space of the spine phantom. The path initiated from one side, traversed to the opposite side, maintained position for several seconds to demonstrate station-keeping capability, executed a directional flip, and returned to the origin. The trajectory was positioned closer to the posterior boundary called the ligamentum flavum (right side in Fig.~\ref{fig:med_workspace}(b)), which comprises a 3-4 mm layer of relatively robust tissue~\cite{chokshi2010thickened}, while maintaining safe distance from the anterior boundary (left side in Fig.~\ref{fig:med_workspace}(b)) adjacent to the spinal cord.

As illustrated in Fig.~\ref{fig:results_med}, the desired trajectory (green line) and actual trajectory (red line) are shown within the epidural space boundaries (purple line). The drug delivery robot achieved a RMS position error of 1.18 mm and orientation error of \SI{7.6}{\degree}. The robot successfully completed the delivery task while maintaining clearance from the anterior boundary (right side in Fig.~\ref{fig:results_med}) throughout the procedure. This experiment demonstrates the potential clinical applicability of the proposed control framework.

\section{Discussion and Conclusion}\label{sec:DiscussionConclusion}
This work addresses an obstacle for clinical magnetic manipulation: maintaining reliable control when feedback is noisy and at low frame rates, as commonly seen in fluoroscopy. We introduced a Zernike polynomial-based model that provides smooth, analytically differentiable representations of the magnetic field and its gradient across the workspace. The proposed control framework based on Zernike model with NMPC and Kalman filter, demonstrates superior performance compared to all baselines and achieves high control accuracy despite low-rate, noisy tracking feedback. Experimental validation in spine phantom studies yielded RMS position errors of 1.18 mm and orientation errors of \SI{7.6}{\degree}, demonstrating the framework's potential for precise drug delivery within spinal anatomy.

There are numerous future research directions to improve clinical viability. First, both modeling and control are restricted to planar motion here. Extending the Zernike representation to volumetric fields and coupling it with a 6-DOF NMPC is an important next step for 3D procedures. Second, the current hardware is heavy and not readily compatible with the C-arm machine. Future work will reduce the system's weight with new frames and integrate true fluoroscopic tracking for ex-vivo and ultimately in-vivo validation. Third, the motion model neglects hydrodynamics and boundary interactions. Incorporating cerebrospinal fluid-like flow and compliant contact with epidural tissues into the estimator and NMPC constraints should improve accuracy and safety. Finally, the spine phantom lacks features related to specific pathology. Adding conditions such as spinal stenosis, ligamentum flavum hypertrophy, disc herniation, or variable epidural fat would better stress the controller and increase clinical relevance.


\bibliography{IEEEabrv,references}
\bibliographystyle{IEEEtran}

\end{document}